\documentclass[twoside,twocolumn,10pt]{article}

\usepackage{wscg}           
\RequirePackage{ifpdf}
\ifpdf
 \RequirePackage[pdftex]{graphicx}
 \RequirePackage[pdftex]{color}
\else
 \RequirePackage[dvips,draft]{graphicx}
 \RequirePackage[dvips]{color}
\fi
\usepackage{booktabs}      
\usepackage{amsmath}        
\usepackage{array}
\newcolumntype{C}[1]{>{\centering\arraybackslash}p{#1}} 

\usepackage{nopageno}       

\usepackage[switch]{lineno}

\title{MVP-Net: Multiple View Pointwise Semantic Segmentation of Large-Scale Point Clouds}

\author{
\parbox{0.25\textwidth}{\centering
Chuanyu Luo\\[1mm]
LiangDao GmbH\\
Germaniastrasse 18-20\\
Germany 12099, Berlin\\[1mm]
chuanyu.luo@liangdao.de
}
\hspace{0.05\textwidth}
\parbox{0.25\textwidth}{\centering
Xiaohan Li, Nuo Cheng, Han Li, Shengguang Lei\\[1mm]
LiangDao GmbH\\
Germaniastrasse 18-20\\
Germany 12099, Berlin\\[1mm]
xiaohan.li, nuo.cheng, han.li, shengguang.lei@liangdao.de
}
\hspace{0.05\textwidth}
\parbox{0.25\textwidth}{\centering
Pu Li\\[1mm]
Ilmenau University of Technology\\
Ehrenbergstrasse 29\\
Germany 98693, Ilmenau\\[1mm]
pu.li@tu-ilmenau.de
}
}

\usepackage{url}
\makeatletter
\def\Uslash{\mathbin{\mathchar`\/}\@ifnextchar{/}{\kern-.15em}{}}
\g@addto@macro\UrlSpecials{\do \/ {\Uslash}}
\def\Ucolon{\mathbin{\mathchar`:}\@ifnextchar{/}{\kern-.1em}{}}
\g@addto@macro\UrlSpecials{\do : {\Ucolon}}
\makeatother

\begin{document}

\twocolumn[{\csname @twocolumnfalse\endcsname

\maketitle  

\begin{abstract}
\noindent
Semantic segmentation of 3D point cloud is an essential task for autonomous driving environment perception. The pipeline of most pointwise point cloud semantic segmentation methods includes points sampling, neighbor searching, feature aggregation, and classification. Neighbor searching method like K-nearest neighbors algorithm, KNN, has been widely applied. However, the complexity of KNN is always a bottleneck of efficiency. In this paper, we propose an end-to-end neural architecture, \textbf{M}ultiple \textbf{V}iew \textbf{P}ointwise Net, MVP-Net, to efficiently and directly infer large-scale outdoor point cloud without KNN or any complex pre/postprocessing. Instead, assumption-based space filling curves and multi-rotation of point cloud methods are introduced to point feature aggregation and receptive field expanding. Numerical experiments show that the proposed MVP-Net is 11 times faster than the most efficient pointwise semantic segmentation method RandLA-Net~\cite{randLA} and achieves the same accuracy on the large-scale benchmark SemanticKITTI dataset.

\end{abstract}

\subsection*{Keywords}
Point Cloud, Semantic Segmentation, Space Filling Curves, Convolutional Neural Networks

\vspace*{1.0\baselineskip}
}]


\section{Introduction}

\copyrightspace

Lidar is widely used in autonomous driving perception systems. The 3D point cloud captured from Lidar provides important geometric information for complex environment perception tasks like object detection and semantic segmentation.

Unlike the regular structured images in computer vision, point cloud is irregular and unordered, and the outdoor large-scale point cloud is sparse. To overcome these challenges, most researchers transform the irregular point cloud to regular projection-based images or 3D voxels. Although these approaches can achieve satisfactory results, there is information loss in 3D-2D projection-based methods. In voxelization methods, the preprocessing is expensive and the computational and memory cost increases cubically by the increase of resolution~\cite{pointCloudSurvey}. In addition, when applying small resolution and large voxel size, the specific granular point features are ignored.

The PointNet~\cite{pointNet} is a pioneering pointwise network that directly processes point cloud and outputs the label per point. The key contribution of PointNet is to introduce the single symmetric
function, max pooling, to aggregate the global features from unordered point clouds. However, the PointNet is limited to small point clouds and cannot be extended to large-scale point clouds~\cite{randLA}. The reason is the learned global features by PointNet cannot represent a large-scale point cloud consisting of many objects and complex structures.

Recently, PointNet-based works~\cite{pointNet2, randLA} were proposed to directly process large-scale point clouds. These pipelines include multi-level point cloud sampling, neighbor searching, and PointNet-based local feature aggregation. RandLA-Net~\cite{randLA} achieves enhanced performances by efficient point cloud random sampling. However, the pointwise methods do not have explicit point neighboring information, and the time-consuming methods like KNN and ball query~\cite{pointCloudSurvey} are still applied to neighbor searching.

\begin{figure}
		\centering 
		\includegraphics[width=0.5\textwidth]{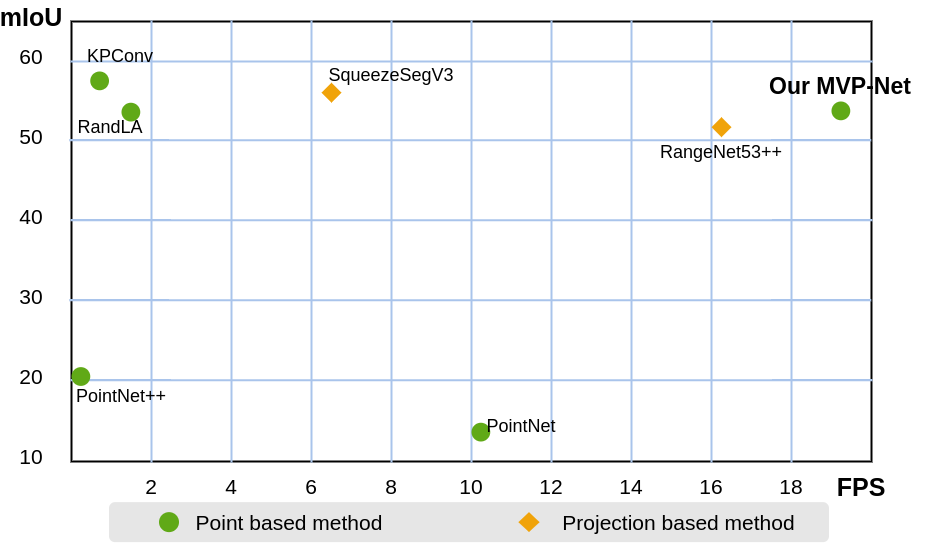} 
		\caption{Quantitative comparison results of our approach and other works on the SemanticKITTI dataset.} 
		\label{fig:comparision} 
\end{figure}

For computer vision semantic segmentation tasks, features aggregation of pixel does not require any extra neighbor searching step, and it is replaced by applying a 2D convolutional layer. The reason for this simplicity is due to the fact that the structure of an image is regular. The most simple strategy to make point cloud regular is to sort the points, and space filling curves is one sorting method to map high dimensional data to 1D sequence, while preserving the 3D local neighbor information. The preserved local information can be further applied to points feature aggregation and prediction. Details of space filling curves can be found in section \ref{sec:sfc}.

\begin{table}
	\centering
	\begin{tabular}{l  c }
	\toprule
	Layer Type & Complexity per Layer \\
	\midrule
	1D Convolution & $O(M \cdot K \cdot C_{in} \cdot C_{out})$\\
	2D Convolution & $O(M^2 \cdot K^2 \cdot C_{in} \cdot C_{out})$\\
	3D Convolution & $O(M^3 \cdot K^3 \cdot C_{in} \cdot C_{out})$\\
	\bottomrule
	\end{tabular}
\caption{Convolutional layer type and complexity per layer. $M$ is the output feature map size. $K$ is the kernel size, $C_{in}$ is the input channel size, and $C_{out}$ is the output channel size. The output feature map size $M$ can be different between different layer type. The 1D convolution layer could be more efficient because it reduces the complexity caused by kernel size $K$} 
\label{table:complexity_per_layer}
\end{table}

Therefore, our basic idea is the 3D Lidar points can be mapped to 1D sequence data, and the proposed 1D convolutional neural network can predict the points by the preserved points local features. As high dimensional neighbor searching method like KNN is deprecated, and high efficient 1D convolution layer is used, our proposed model is more efficient than the other similar benchmark methods by a great margin (see Fig.\ref{fig:comparision}). The efficiency of 1D convolution layer and  the complexity comparison of other layers can be found in Tab.\ref{table:complexity_per_layer}.

\section{Related Work}
\label{sec:relatedWork}

In this section, we give an overview of deep learning based approaches in point cloud semantic segmentation tasks, mainly including voxelization based methods, projection based methods, and point based methods. Besides, the sorting algorithm space filling curves (SFC) will also be introduced.

\subsection{Voxelization based methods}
Voxelization based methods~\cite{Cylinder3D,PolarNet,RPVNet, SPVNAS} transform the irregular unordered point cloud into regular 3D grids, and then the powerful 3D convolution is applied in feature extraction and prediction. However, the problem of granular information loss can be caused by using a large voxel size. Meanwhile, the computational and memory cost increases cubically by the increase of the resolution, especially when processing large-scale outdoor sparse point cloud~\cite{pointCloudSurvey}. To tackle the granular information loss caused by a large voxel size and the expensive cost caused by a small voxel size, the fusion of a point based method and sparse 3D convolution can be a feasible solution.

\subsection{Projection based methods}
To leverage the success from 2D image processing neural networks, the projection based methods~\cite{squeezesegv3} project the 3D point cloud into 2D images, and then the traditional 2D architecture like U-net~\cite{unet} is applied for features aggregation. However, the primary limitation lies in the information loss from 3D-2D projection~\cite{pointCloudSurvey}.

\subsection{Point based methods}
Inspired by the pioneering work PointNet, many works~\cite{pointNet2, randLA} were introduced by directly taking the point cloud as input and learning the pointwise features by multi-layer perception and symmetric function like max pooling.

To directly process large-scale outdoor point cloud, in most pointwise MLP based works~\cite{pointNet2, randLA} the pipeline includes sampling, neighbor points searching such as ball query or K-nearest-neighbors (KNN), PointNet-based features extraction, and per-point classification. Recently, RandLA-Net\cite{randLA} was developed to utilize random point sampling and showed impressive results. However, for carrying out the neighbor points searching task, RandLA-Net still utilizes the KNN method, which limits its efficiency even using the KDTree~\cite{kdtee} algorithm with complexity $O(nlog(n))$.

KPConv\cite{kpconv} is a kernel based point convolution method, which in essence takes the neighbor points as input and processes the points with spatial kernel weights. However, in this method, the radius neighborhood approach is applied to neighbor searching, and the network cannot directly train the entire data of large scenes~\cite{RPVNet}.

\subsection{Space filling curves}

\label{sec:sfc}

\begin{figure}
\centering 
\includegraphics[width=0.4\textwidth]{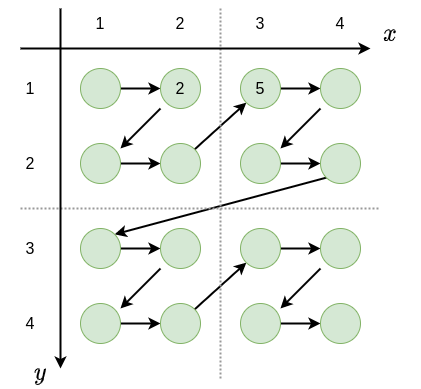} 
\caption{Morton-order in 2D case. The 2D points is sorted while the locality is preserved, but there is still information loss, like the distance between point 2 and 5 in 2D and 1D case is different.} 
\label{fig::morton_code} 
\end{figure}

Space filling curves (SFC) is a sorting method to map high dimensional data to one dimension sequence, while preserving locality of the data points (e.g.\ the Euclidean neighbor similarity in 3D tends to be kept after mapped to 1D)~\cite{MortonNet}. One of widely applied and high efficient SFC methods is Morton-order~\cite{MortonOrder}, also known as Z-order because of the shape of the curve in the 2D case, e.g.\ the curve in Fig.\ref{fig::morton_code}. The other SFC methods~\cite{SFC_intro} include Hilbert, Peano, Sierpinski curves, etc. 

In point cloud field, MortonNet~\cite{MortonNet} uses Morton-order in self-supervised tasks, which not require semantic labels. The trained MortonNet predicts the next point in a point sequence created by Morton-order, and the results show the learned Morton features can be transferred to semantic segmentation tasks and improve performances. However, like PointNet, MortonNet is limited to small-scale point clouds.

Though SFC can keep the local features of high dimensional data after mapping, there is still information loss. As shown in Fig.\ref{fig::morton_code}, the distance between point 2 and point 5 in 2D space is $1$, but after mapping to 1D sequence, the distance in 1D space is $3$. The point 5 might not be considered as the neighbor of point 2 after mapping, which limits the network points feature aggregation and prediction ability.


\section{MVP-Net}
\label{sec:mvp}

In this section, we first present our key contribution, the assumption-based space filling curves for 3D point cloud, and the expansion of receptive field per point by rotating the raw points, and at last an overview of our network architecture MVP-Net, \textbf{M}ultiple \textbf{V}iew \textbf{P}ointwise Net, and the implementation details.


\subsection{Space filling curves for 3D point cloud}

Space filling curves (SFC) are widely applied to the regular dense data like images, matrices and grid cells~\cite{SFC_intro}. To obtain the SFC of large-scale irregular sparse point cloud, we propose the Eq.\ref{eq:sorting} to calculate the score per point, and sort the points order by the calculated scores.

\begin{figure}
\centering 
\includegraphics[width=0.5\textwidth]{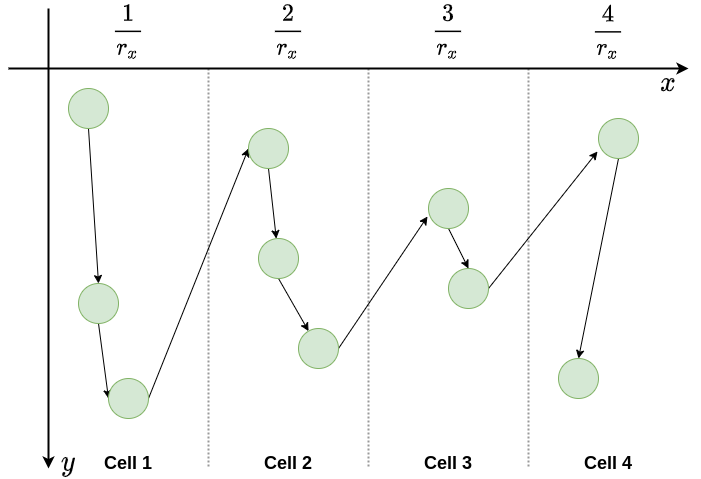} 
\caption{Space filling curves for 2D point cloud by simple scores Eq.\ref{eq:sorting_simple}} 
\label{fig::simple_sorting} 
\end{figure}

For illustration purpose, the most simple case of 2D points and scores per point is expressed as Eq.\ref{eq:sorting_simple}, 
\begin{equation}
\begin{split}
  scores = k_x \cdot round(x \cdot r_x) + y
\end{split}
\label{eq:sorting_simple}
\end{equation}

where $x$, $y$ denote the coordinates of 2D point,  $round()$ is the rounding function to find the nearest integer of the input. The key goal is to sort the points by cells along $x$ axis, and in each cell the points will be sorted along $y$ axis, as shown in Fig.\ref{fig::simple_sorting}. The points in Fig.\ref{fig::simple_sorting} will be sorted by the scores in Eq.\ref{eq:sorting_simple} in ascending order. $r_x$ and $k_x$ are two hyperparameters to make the points strictly sorted. 

The cell width is $\frac{1}{r_x}$, it can be derivated by inequation \ref{eq:sorting_derivation_1} and \ref{eq:sorting_derivation_2}

\begin{equation}
-0.5 \leq x \cdot r_x - x_R  < 0.5
\label{eq:sorting_derivation_1}
\end{equation}
\begin{equation}
\rightarrow  \quad  \frac{x_R}{r_x} - \frac{0.5}{r_x} \leq x  < \frac{x_R}{r_x} + \frac{0.5}{r_x}
\label{eq:sorting_derivation_2}
\end{equation}

where $x_R$ is the nearest integer of  $x \cdot r_x$, i.e. $x_R = round(x \cdot r_x)$.

To make sure the points are sorted strictly cell by cell, the condition that the last point score in cell $i$ is less than the first point score in cell $i + 1$ should be always satisfied.

If the last point coordinate in cell $i$ is denoted as $(x_i, y_i)$, and the first point in next cell $i + 1$ is $(x_{i+1}, y_{i+1})$, the rounded values along $x$ can be denoted as $x_R^i = round(x_i \cdot r_x)$ and $x_R^{i+1} = round(x_{i+1} \cdot r_x)$. Since cell width is $\frac{1}{r_x}$, the rounded values satisfy $x_R^{i+1} - x_R^i = \frac{1}{r_x}$.

To make sure the last point scores in cell $i$ should be less than the first point scores in cell $i + 1$, also $k_x \cdot x_R^i + y_i < k_x \cdot x_R^{i+1} + y_{i+1}$, the hyperparameter $k_x$ should satisfy

\begin{figure*}[h!]
\centering 
\includegraphics[width=0.96\textwidth]{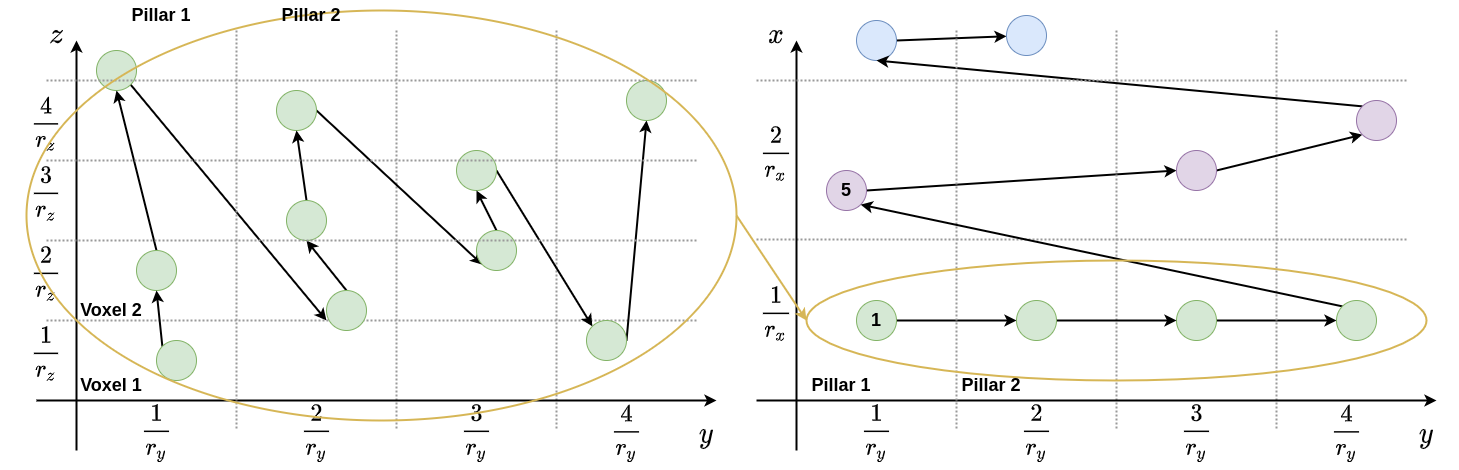} 
\caption{Space filling curves for 3D point cloud by scores Eq.\ref{eq:sorting}} 
\label{fig:full_case} 
\end{figure*}

\begin{equation}
k_x \cdot x_R^i + y_i < k_x \cdot x_R^{i+1} + y_{i+1}
\label{eq:sorting_derivation_3}
\end{equation}
\begin{equation}
\rightarrow  \quad  k_x >  \frac{y_{i} - y_{i+1}}{x_R^{i+1} - x_R^{i}} = (y_{i} - y_{i+1}) \cdot r_x
\label{eq:sorting_derivation_4}
\end{equation}
\begin{equation}
\rightarrow  \quad  k_x > (y_{max} - y_{min}) \cdot r_x,
\label{eq:sorting_derivation_5}
\end{equation}

where $y_{max}$ and $y_{min}$ is maximal and minimal coordinate $y$ values in all point clouds. In practice, we focus on the region of interest points, e.g.\ in a rectangular region of $X_{min} < x < X_{max}$, and $Y_{min} < y < Y_{max}$. As a result, the hyperparameter $ k_x$ should meet the condition \ref{eq:sorting_derivation_6}

\begin{equation}
k_x > (Y_{max} - Y_{min}) \cdot r_x.
\label{eq:sorting_derivation_6}
\end{equation}

The hyperparameter $r_x$, which determines the cell size, will be extended and discussed in the full sorting equation.

Based on Eq.\ref{eq:sorting_simple}, Fig.\ref{fig::simple_sorting} shows that the points will be sorted at first along $y$ axis, and then cell by cell along $x$ axis, as the $x$ item will contribute more scores after multiplying $k_x$.

To introduce the complete space filling curves of 3D point cloud in an autonomous driving scene, at first we define two assumptions to aggregate the most important point cloud neighboring features for prediction,
\begin{enumerate}
\item For neighbor-points searching, the neighbor per point should be the same object as the point to be classified as much as possible. 
\item In an outdoor autonomous driving scene, the points, except ground points, along object height $z$ axis are more possible to belong to the same object than those along the other direction.
\end{enumerate}

The first assumption can be explained by an example that a point is classified as a car is based on the information of the other points on the same car, instead of the other object points, even when the other object points are more spatially close to the to be classified point.
The second assumption can be explained by standing object examples like cars, trees, buildings, etc. Except for standing object points, the ground points have distinguishing features and thus can be easily predicted.

From the two assumptions above, we extend Eq.\ref{eq:sorting_simple} to Eq.\ref{eq:sorting}, in which  the feature along the object height axis is more important, and the points along $z$ axis in each pillar will be at first aggregated.

\begin{equation}
\begin{split}
  scores = k_x \cdot round(x \cdot r_x) + k_y \cdot round(y \cdot r_y) \\ + k_z  \cdot round(z \cdot r_z) + k_{\rho} \cdot {\rho}
\end{split}
\label{eq:sorting}
\end{equation}

where
\begin{equation}
\begin{split}
 \rho = \sqrt{x^2 + y^2}
\end{split}
\end{equation}

where $x$, $y$, $z$ denote the coordinates of each point, $ k_x \gg k_y \gg k_z \gg k_{\rho}$. The hyperparameter $r_x$ and $r_y$ determine the pillar size. Small pillar size will only contain few points in each pillar, and lose feature information. Large pillar size is against the first assumption, because the pillar will include different object points. These parameters, including the pillar size, are set empirically in implementation as Tab.\ref{table:parameters}.

From the discussion in the simplified sorting Eq.\ref{eq:sorting_simple}, the 3D points will be sorted at first along $z$ axis voxel by voxel, and then along $y$ axis and at last along $x$ axis pillar by pillar. The space filling curve, as shown in Fig.\ref{fig:full_case}, derivated by Eq.\ref{eq:sorting}, preserve the most useful local features after mapping from 3D data to 1D sequence.

The complexity of KNN, applied in 3D case, is $O(nlog(n))$ when applying the KDTree~\cite{kdtee} algorithm, but the complexity of the proposed sorting function Eq.\ref{eq:sorting} is only $O(n)$.

\begin{table}
	\centering
	\begin{tabular}{l  c }
	\toprule
	Parameter & value \\
	\midrule
	$k_x$ & $10^{10}$\\
	$k_y$ & $10^5$\\
	$k_z$ & $10^0$\\
	$k_{\rho}$ & $10^{-5}$\\
	\midrule
	$r_x$ & $1.2$\\
	$r_y$ & $1.2$\\
	$r_z$ & $4$\\
	\bottomrule
	\end{tabular}
\caption{The parameters and values of sorting Eq.\ref{eq:sorting} in implementation} 
\label{table:parameters}
\end{table}


\begin{figure*}[h!]
\centering 
\includegraphics[width=0.96\textwidth]{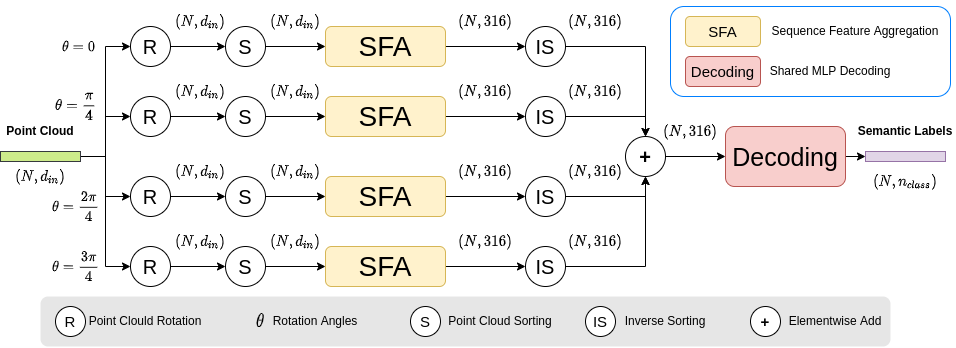} 
\caption{The architecture of the MVP-Net. $N$ represents the number of points. $d_{in}$ denotes the input features of point including the coordinates and other features. $n_{class}$ is the number of classes to be predicted. The point cloud rotation expands the receptive field per point. The point sorting makes the point sequence regular, and SFA can directly aggregate the neighbor information. After inversely sorted to the original order, the aggregated features will be decoded by a share-MLP based network.} 
\label{fig:architecture} 
\end{figure*}

\subsection{Point cloud rotation}

Though the assumption-based sorting Eq.\ref{eq:sorting} keeps the most meaningful features along the height $z$ axis, there is still information loss caused by SFC, which is introduced in section \ref{sec:sfc}. In Fig.\ref{fig:full_case}, e.g.\ the pillar points 5  is the neighbor of pillar points 1 in 3D space, but the distance is much longer in 1D space. Therefore, the receptive field per point in 1D is limited. To solve this problem, the raw point cloud will be rotated along $z$ axis multiple times in our work, and the idea is similar to the multiple view in projection methods.

If sorting scores Eq.\ref{eq:sorting} is applied to the rotated by angle $\frac{\pi}{2}$ point cloud, it is equivalent applying Eq.\ref{eq:sorting_rotated}, where $ k_y \gg k_x \gg k_z \gg k_{\rho}$ and the direction along $x$ and $y$ axis is exchanged compared with Eq.\ref{eq:sorting}, to the unrotated point cloud. In this case, the pillar points 5 is the neighbor of pillar 1. In our proposed network, the raw point cloud will be rotated by 4 different angles, and by applying only Eq.\ref{eq:sorting}, the neighbor and receptive field per point will be expanded.

\begin{equation}
\begin{split}
  scores = k_y \cdot round(y \cdot r_y) + k_x \cdot round(x \cdot r_x) + \\ k_z  \cdot round(z \cdot r_z) + k_{\rho} \cdot {\rho}  
\end{split}
\label{eq:sorting_rotated}
\end{equation}



\subsection{Architecture}

The architecture of MVP-Net is illustrated in Fig.\ref{fig:architecture}. The raw point cloud is at first rotated along the $z$ axis by 4 fixed angles, $0$, $\frac{\pi}{4}$, $\frac{2\pi}{4}$, $\frac{3\pi}{4}$. The rotated point clouds are then sorted by Eq.\ref{eq:sorting}. After being sorted, the point cloud is regular and has explicit neighbor information.

\begin{figure}
		\centering 
		\includegraphics[width=0.5\textwidth]{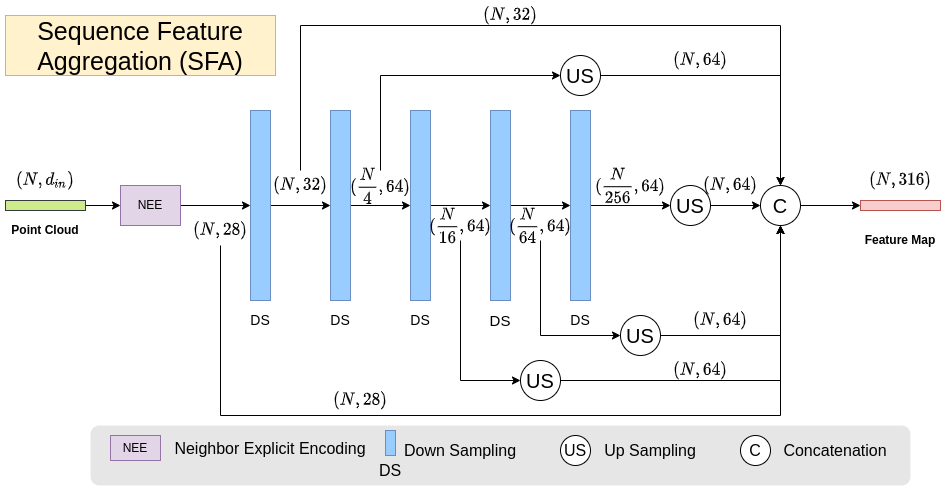} 
		\caption{Sequence Feature Aggregation (SFA). SFA is a network consisting of only 1D convolution layer for regular sequence point cloud feature aggregation.} 
		\label{fig:SFA} 
\end{figure}

The sequence feature aggregation backbone SFA, \textbf{S}equence \textbf{F}eature \textbf{A}ggregation, as shown in Fig.\ref{fig:SFA}, is a 1D convolutional network to aggregate the local and global features per point.

For each point $i$, the position difference between $p_i$ and its nearest 8 points $\{p_i^1 ... p_i^k ... p_i^8\}$ along the sorted sequence are explicitly encoded to achieve the translation invariance~\cite{wu2019pointconv} of the point cloud as follows:

\begin{equation}
\begin{split}
  x_i = p_i \oplus (p_i - p_i^k) \oplus p_i^{other \ features}
\end{split}
\label{eq:nee}
\end{equation}
where $p_i$ and $p_i^k$ are the $xyz$ coordinates of the point $i$ and the neighbor point. $p_i^{other \ features}$ represents the other features except $xyz$ positions of the point $i$, e.g.\ the intensity of reflection per point. $\oplus$ is the concatenation operation. The features extracted by SFA will be inversely sorted by the sorting function to the original order. The inverse sorted features per point from different angles will be added and then decoded by a multi-layer perception based network with a kernel size of $1$.

\begin{figure*}[h!]
\centering 
\includegraphics[width=0.98\textwidth]{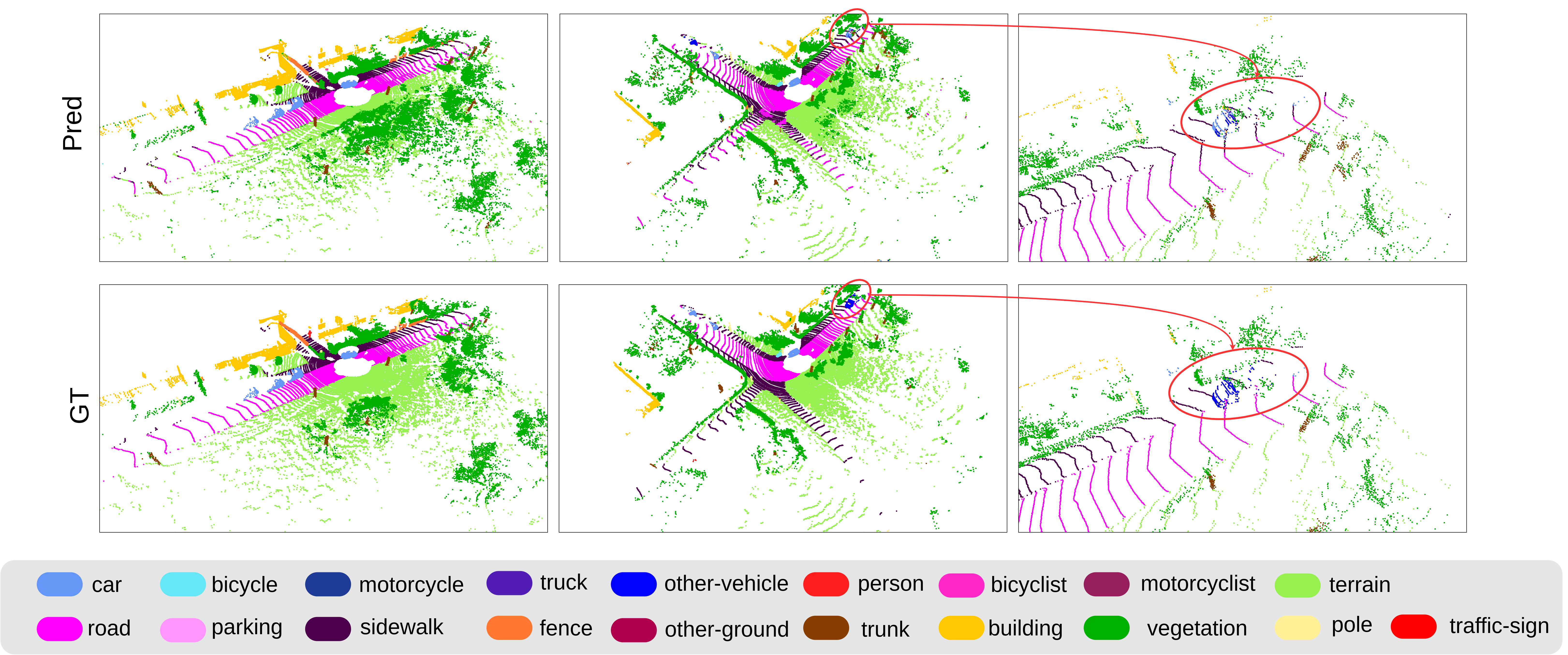} 
\caption{Qualitative comparison of prediction and ground truth on the validation set. The red circle shows the failure case of other-vehicle points classification.} 
\label{fig:visualization} 
\end{figure*}

\subsection{Implementation}

We implement our network in Pytorch~\cite{pytorch}. For batch training, we sample the input cloud to $10^5$ points per frame. We use the Adam optimizer~\cite{adam} with default parameters and the learning rate is set as $0.0003$ without any decay. The most commonly used cross-entropy loss is employed as the loss function at first. When the network optimization steps into a plateau, we adapt the loss function as
\begin{equation}
\begin{split}
 \text{Loss} = \text{Cross-entropy-loss + Lov\'asz-softmax-loss}
\end{split}
\end{equation}

The Lov\'asz-Softmax-loss~\cite{lovasz} is a differentiable loss function designed to maximize the mean intersection-over-union (mIoU) score directly, which is commonly employed in the evaluation of semantic segmentation tasks. For data augmentation, we randomly rotate the original point cloud along $z$ axis at each training step. For validation and testing, we input the whole original point cloud into the network and infer the semantic labels per point without any pre/postprocessing. All experiments are conducted on a Tesla V100 with one GPU 32G.

\section{Experiments}
\label{sec:exp}
\subsection{Dataset}
We trained and tested our MVP-Net on the SemanticKITTI~\cite{semanticKitti} dataset which provides large-scale semantic annotation per point in autonomous driving scenes. The point annotations for individual scan of the sequence 00-10 are provided. We used sequence 08 as a validation set and the other 10 sequences as a training set. Each point includes the information of the three dimensional coordinates and the remission. The mean intersection-over-union (mIoU) over 19 classes is used as the standard metric.

\subsection{Results}
Fig.\ref{fig:visualization} presents a qualitative result of the prediction and the ground truth on the validation set.

\begin{table*}
  \centering
  \scriptsize
  
  \begin{tabular}{p{2.2cm} C{0.3cm} C{0.3cm} C{0.18cm} C{0.18cm} C{0.18cm} C{0.18cm} C{0.18cm} C{0.18cm} C{0.18cm} C{0.18cm} C{0.18cm} C{0.18cm} C{0.18cm} C{0.18cm} C{0.18cm} C{0.18cm} C{0.18cm} C{0.18cm} C{0.18cm} C{0.18cm} C{0.18cm} C{0.18cm}}
    \toprule
    Method & FPS & \rotatebox{90}{mIoU($\%$)} & \rotatebox{90}{road} & \rotatebox{90}{sidewalk} & \rotatebox{90}{parking} & \rotatebox{90}{other-ground} & \rotatebox{90}{building} & \rotatebox{90}{car}  & \rotatebox{90}{truck} & \rotatebox{90}{bicycle} & \rotatebox{90}{motorcycle} &\rotatebox{90}{other-vehicle} & \rotatebox{90}{vegetation} & \rotatebox{90}{trunk} & \rotatebox{90}{terrain} & \rotatebox{90}{person} & \rotatebox{90}{bicyclist} & \rotatebox{90}{motorcyclist} & \rotatebox{90}{fence} & \rotatebox{90}{pole} & \rotatebox{90}{traffic-sign}\\
    \midrule
    PointNet~\cite{pointNet} & 10.12 & 14.6 &  61.6 & 35.7 &  15.8 & 1.4 & 41.4 & 46.3 & 0.1 & 1.3 & 0.3 & 0.8 & 31.0 & 4.6 & 17.6 & 0.2 & 0.2 & 0.0 & 12.9 & 2.4 & 3.7 \\
    PointNet++~\cite{pointNet2} & 0.06 & 20.1 &  72.0 & 41.8 &   18.7 & 5.6 & 62.3 & 53.7 & 0.9 &  1.9 &  0.2 &  0.2 & 46.5 &  13.8 &  30.0 & 0.9 &  1.0 & 0.0 & 16.9 & 6.0 & 8.9 \\
    RandLA~\cite{randLA} & 1.74 & 53.9 &  90.7 & 73.7 &   60.3 & 20.4 &  86.9 & 94.2 & 40.1 &  26.0 &   25.8 &  38.9 &  81.4 &  61.3 &  66.8 & 49.2 &  48.2 & 7.2 & 56.3 & 49.2 & 47.7 \\
    KPConv~\cite{kpconv} & 0.88 & 58.8 &  88.8 & 72.7 &    61.3  & 31.6 & 90.5 & 96.0 & 33.4 &  30.2 & 42.5 &  44.3 &  84.8 &  69.2 &  69.1 & 61.5 &  61.6 & 11.8 & 64.2 & 56.4 & 47.4 \\
    \midrule
    
    SqueezeSegV3~\cite{squeezesegv3} & 6.49 & 55.9 &  91.7 & 74.8 & 63.4 & 26.4 &  89.0 & 92.5 & 29.6 &  38.7 & 36.5 & 33.0 &  82.0 &   59.4 &   65.4 & 45.6 &  46.2 & 20.1 & 58.7 & 49.6 &  58.9 \\
    
    RangeNet53++~\cite{RangeNet++} & 16.12 & 52.2 & 91.8 & 75.2 & 65.0 & 27.8 & 87.4 &  91.4 & 25.7 & 25.7 & 34.4 &  23.0 & 80.5 & 55.1 &  64.6 & 38.3 & 38.8 &  4.8 &  58.6 & 47.9 & 55.9 \\
    \midrule
    Cylinder3D~\cite{Cylinder3D} & 2.55 & 67.8 & 91.4 & 75.5 & 65.1 & 32.3 & 91.0 & 97.1 & 59.0 & 67.6 & 64.0 & 58.6 & 85.4 & 71.8 & 71.8 & 73.9 & 67.9 & 36.0 & 66.5 & 62.6 & 65.6\\
    PolarNet~\cite{PolarNet} & 1.96 & 54.3 &  90.8 & 74.4 &   61.7 & 21.7 & 90.0 & 93.8 & 22.9 &  40.3 &    30.1 &  28.5 &  84.0 &  61.3 &   65.5 & 43.2 &  40.2 & 5.6 & 61.3 & 51.8 & 57.5 \\
   
    \midrule
     SPVNAS~\cite{SPVNAS} & 0.88 & 66.4 & - & - & - & - & - & - & - & - & - & - & - & - & - & - & - & - & - & - & - \\
     RPVNet~\cite{RPVNet} & - & \textbf{70.3} & 93.4 & 80.7 & 70.3 & 33.3 & 93.5 & 97.6 & 44.2 & 68.4 & 68.7 & 61.1 &  86.5 & 75.1 & 71.7 & 75.9 & 74.4 &  73.4 & 72.1 & 64.8 & 61.4\\
    \midrule
    MVP-Net & \textbf{19.20} & 53.9 &  91.4 & 75.9 & 61.4 & 25.6 &  85.8 & 92.7 & 20.2 &  37.2 & 17.7 & 13.8 &  83.2 &   64.5 & 69.3 & 50.0 &  55.8 & 12.9 & 55.2 & 51.8 & 59.2 \\
    \bottomrule
  \end{tabular}
  \caption{The comparison results of our network and the other recently published point-based, projection-based, voxelization-based and fusion-based methods on SemanticKITTI dataset. The input of points is fixed as $10^5$, and the size of projection image and voxel is set the same as literature. FPS, frames per second, represents the inference speed. mIoU, intersection-over-union, is the accuracy metric over all classes.}
  \label{tab:example}
\end{table*}

Tab.\ref{tab:example} presents a quantitative comparison of our MVP-Net with the other recently published methods. These methods are grouped by point-based~\cite{pointNet, pointNet2, randLA, kpconv}, projection-based~\cite{squeezesegv3, RangeNet++}, voxelization-based~\cite{Cylinder3D, PolarNet}, and fusion-based methods~\cite{SPVNAS, RPVNet}.

The comparison demonstrates that the voxelization-based methods surpass the point-based and projection-based methods on accuracy with large margin. The fusion-based method RPVNet~\cite{RPVNet}, the fusion of point, projection and voxelization method, ranks 1st of accuracy without much surprise.

Our proposed MVP-Net achieves the highest efficiency over all the other methods. As one point-based method, MVP-Net also achieve the comparable accuracy, and the same mIoU as the pointwise benchmark method, RandLA-Net~\cite{randLA}, which uses KNN searching for point neighbor in 3D space. This result shows that the 3D points local feature can be preserved by space filling curves. The general information loss problem is solved by our assumption-based curves and point cloud rotation. The proposed contributions will also be proved in ablation study section \ref{sec:ablation}.

\subsection{Ablation Study}
\label{sec:ablation}

All the ablated networks are trained by the sequence 00-07 and 09-10, and evaluated by the sequence 08 of the SemanticKITTI dataset.
\begin{enumerate}

\item \textbf{Increasing/decreasing point cloud rotation times}. To achieve an excellent trade-off between efficiency and effectiveness, the number of point cloud rotation along $z$ axis is studied.

\item \textbf{Removing the point neighbor explicit encoding (NEE) in the SFA unit}. To achieve the translation-invariance of the point cloud, the $xyz$ positions difference of the center point and its neighbor point are explicitly encoded in the SFA unit. The point neighbor explicit encoding step is removed in the ablation study, and the original point features are directly fed into the SFA unit.

\item \textbf{Replace the assumption-based sorting function by another function}. The assumption based sorting function is one key contribution in our study, and we think the feature along object height axis is more important. Here we compare the original sorting Eq.\ref{eq:sorting} with Eq.\ref{eq:alaternative}, which will not preferentially aggregate the points along $z$ axis.
\end{enumerate}

\begin{table}
	\centering
	\begin{tabular}{l  c  c }
	\toprule
	Ablation Experiment & FPS & mIoU \\
	\midrule
	Rotating the point cloud once & 55.2 & 47.4 \\
	Rotating the point cloud twice & 33.2 & 51.7 \\
	Rotating the point cloud eight times & 10.5 & 54.8 \\
	\midrule
	Removing the NEE in SFA unit & 19.8 & 50.2 \\
	\midrule
	Sorting by Eq.\ref{eq:alaternative} & 19.2 & 43.7\\
	\midrule
	Original framework & 19.2 & 54.6 \\
	\bottomrule
	\end{tabular}
	
\caption{Ablation experiments based on the original framework}
\label{tab:ablation}

\end{table}

\begin{equation}
\begin{split}
 scores = k_z  \cdot round(z \cdot r_z) + k_x \cdot round(x \cdot r_x) \\ + k_y \cdot round(y \cdot r_y) +  k_{\rho} \cdot {\rho}
\end{split}
\label{eq:alaternative}
\end{equation}
The parameters of Eq.\ref{eq:alaternative} meet $ k_z \gg k_x \gg k_y \gg k_{\rho}$, and the other parameters are the same as the original sorting Eq.\ref{eq:sorting}. All results are listed in Tab.\ref{tab:ablation}. It can be seen that the proposed parameters and methods achieve the best trade-off between efficiency and effectiveness. The proposed assumption-based SFC by Eq.\ref{eq:sorting} also surpasses the normal SFC, e.g. Eq.\ref{eq:alaternative}.

\section{Conclusion}

In this paper, we present a novel end-to-end pointwise network MVP-Net for 3D point cloud semantic segmentation tasks. In contrast to the existing pointwise methods, our network shows the possibility to totally remove the KNN or other neighbor searching methods. Instead, assumption-based space filling curves and point cloud rotation from multiple angles are proposed to achieve comparable accuracy and high efficiency. 
 
Our model is a fully 1D convolutional architecture without any pre/postprocessing, and therefore, the state-of-the-art neural modules from computer vision and natural language processing can be directly imported into our model. The fusion with other types of methods like voxelization will also be explored in future work.

As far as we know, our method is the first to directly apply the space filling curves to large-scale point cloud, and our experiments prove that the local information loss problem caused by SFC can be solved in point cloud field. The defined two assumptions are explained and proved by the experiments and ablation study, which is important for rethinking the features aggregation in autonomous driving scene.


\end{document}